\pdfoutput=1

\documentclass{article}

\usepackage[final]{neurips_2024}

\usepackage[utf8]{inputenc} 
\usepackage[T1]{fontenc}    
\usepackage{hyperref}       
\usepackage{url}            
\usepackage{booktabs}       
\usepackage{amsfonts}       
\usepackage{nicefrac}       
\usepackage{microtype}      
\usepackage{xcolor}         

\usepackage{times}
\usepackage{latexsym}
\usepackage{tcolorbox}
\usepackage[T1]{fontenc}

\usepackage[utf8]{inputenc}
\usepackage{comment}

\usepackage{microtype}

\usepackage{inconsolata}

\usepackage{graphicx}

\usepackage{multirow}
\usepackage{xcolor}
\usepackage{mdframed}

\usepackage{enumitem}
\usepackage{amsmath}

\newtcolorbox{insightbox}{
  colback=yellow!10!white, 
  colframe=yellow!50!black, 
  title=Insight, 
  fonttitle=\bfseries, 
  boxrule=0.5mm, 
  arc=4mm, 
  left=1mm, 
  right=1mm, 
  top=1mm, 
  bottom=1mm, 
  width=\linewidth, 
  enhanced, 
}

\makeatletter
\def\UrlAlphabet{%
      \do\a\do\b\do\c\do\d\do\e\do\f\do\g\do\h\do\i\do\j%
      \do\k\do\l\do\m\do\n\do\o\do\p\do\q\do\r\do\s\do\t%
      \do\u\do\v\do\w\do\x\do\y\do\z\do\A\do\B\do\C\do\D%
      \do\E\do\F\do\G\do\H\do\I\do\J\do\K\do\L\do\M\do\N%
      \do\O\do\P\do\Q\do\R\do\S\do\T\do\U\do\V\do\W\do\X%
      \do\Y\do\Z}
\def\UrlDigits{\do\1\do\2\do\3\do\4\do\5\do\6\do\7\do\8\do\9\do\0}
\g@addto@macro{\UrlBreaks}{\UrlOrds}
\g@addto@macro{\UrlBreaks}{\UrlAlphabet}
\g@addto@macro{\UrlBreaks}{\UrlDigits}
\makeatother

\title{Understanding and Alleviating Memory Consumption in RLHF for LLMs}

\author{%
  Jin Zhou, Hanmei Yang, Steven (Jiaxun) Tang, Mingcan Xiang, Hui Guan \\
  University of Massachusetts Amherst, MA, USA \\
  \texttt{\{jinzhou, hanmeiyang, jtang, mingcanxiang, huiguan\}@umass.edu} \\
  \And
  Tongping Liu \\
  ByteDance Inc., San Jose, CA, USA \\
  \texttt{tongping.liu@bytedance.com} \\
}

\begin{document}
\maketitle
\begin{abstract}
Fine-tuning with Reinforcement Learning with Human Feedback (RLHF) is essential for aligning large language models (LLMs). However, RLHF often encounters significant memory challenges. 
This study is the first to examine memory usage in the RLHF context, exploring various memory management strategies and unveiling the reasons behind excessive memory consumption. Additionally, we introduce a simple yet effective approach that substantially reduces the memory required for RLHF fine-tuning.

\end{abstract}

\section{Introduction}
\label{sec:intro}

Reinforcement Learning from Human Feedback (RLHF) helps align large language models (LLMs) with human values and expectations~\cite{InstructGPT,ChatGPT,GPT4}, ensuring that the models produce more accurate, relevant, and contextually appropriate outputs. The RLHF process includes multiple inference and training phases, which involve a total of four models, leading to high memory consumption~\cite{DeepSpeedChatIssue1, ColossalChatIssue, DeepSpeedChatIssue2}. Efficient memory management techniques are necessary to enable the practical deployment of RLHF.

Improving memory efficiency during training and inference is a well-studied topic. Previous studies have proposed different memory management policies to reduce memory consumption, such as Zero Redundancy Optimizers (ZeRO)~\cite{ZeRO12, ZeRO3}, gradient checkpointing~\cite{Checkpointing}, and CPU offloading~\cite{Offload}, as discussed in \S\ref{sec:memorystrategies}. Additionally, some work focuses on reducing the memory consumption of inference, especially for key-value caching~\cite{kwon2023efficient, prabhu2024vattention}. 
These memory management strategies are typically combined together to optimize the memory consumption, as they are orthogonal in theory. However, \textit{based on our experiments, some strategies actually introduce higher memory consumption, instead of reducing it}. 

Therefore, it is crucial to understand the reason behind this, and how each memory management strategy may inadvertently affect the overall memory consumption. This knowledge will enable the development of optimized memory management strategies and impact the cost of computational resources, making the RLHF fine-tuning of LLMs more accessible and sustainable.

For this purpose, this work provides the first study on memory usage in the RLHF scenario. During this study, we focus on two open-source RLHF frameworks, including DeepSpeed-Chat~\cite{DeepSpeedChat} and ColossalChat~\cite{ColossalChat}. Our study also includes two types of LLM models, aiming to answer the following technical questions:

\setlist{nolistsep}
\begin{itemize}[noitemsep]
    \item \textbf{R1:} What causes high memory consumption during RLHF?  
    \item \textbf{R2:} How effective are different memory management strategies? 
    \item \textbf{R3:} How can we effectively reduce memory consumption in RLHF? 
\end{itemize}

Overall, this study explains why enabling certain strategies may introduce counter-intuitive effects in memory reduction, and further proposes a simple yet effective method that helps reduce memory consumption in RLHF training. The proposed approach minimizes memory consumption without requiring substantial code changes or redesigns, thus offering an efficient solution with minimal effort.

\section{Background}
\subsection{RLHF Training}

RLHF~\cite{InstructGPT} is a widely used technique to enhance LLMs by incorporating human preferences into the training process. It involves three stages, each requiring fine-tuning of the LLM: (1) supervised fine-tuning (SFT) on an instruction-following dataset; (2) training a reward model on human preference data; and (3) fine-tuning via proximal policy optimization (PPO)~\cite{schulman2017proximal}. 
Our work focuses on the third stage, which is resource-intensive due to the need to manage several large models: the SFT reference model to prevent reward divergence, the reward model for calculating sequence returns, the actor model (final RLHF-aligned LLM) initialized from the reference model, and the critic model initialized from the reward model to estimate returns.
In this stage, the actor generates responses to prompts, which are then evaluated by multiple inferences (actor, reference, critic, and reward) to produce experience data. Finally, the actor model is trained to maximize rewards while minimizing policy deviation, and the critic model is trained to ensure alignment with human preferences.

\subsection{Memory Management in Training}
\label{sec:memorystrategies}

Various strategies can reduce memory consumption during model training. ZeRO~\cite{ZeRO12,ZeRO3} minimizes data redundancy in distributed training by partitioning optimizer states, gradients, and model parameters. CPU Offloading~\cite{Offload} moves some data to CPU memory, thereby reducing GPU memory consumption. Gradient checkpointing~\cite{Checkpointing} trades computation for reduced memory usage by storing only partial activations and recomputing the rest. Existing RLHF training systems often rely on open-source LLM training frameworks that incorporate these memory optimization techniques. For instance, ColossalChat~\cite{ColossalChat} uses Colossalai~\cite{Colossalai}, while DeepSpeed-Chat~\cite{DeepSpeedChat}, trlX~\cite{trlX}, APP~\cite{APP}, and LLaMA-Factory~\cite{LlamaFactory} use DeepSpeed~\cite{DeepSpeed}.

These training frameworks are typically built on top of PyTorch, relying on its CUDA caching allocator for managing memory usage.
Pytorch allocator exposes two parameters that help to understand memory usage: \textbf{reserved memory} refers to the total amount of GPU memory that has been reserved by the PyTorch CUDA caching allocator from the CUDA driver, and \textbf{allocated memory} refers to the amount of GPU memory that is currently being used by active tensors.

\section{Detailed Studies}
This section develops the experiments to answer the three research questions in \S\ref{sec:intro}, based on two open-source RLHF projects. In the remainder of this section, the size of fragmentation is equal to the difference between reserved memory and allocated memory when the PyTorch allocator attempts to allocate more memory from the CUDA driver.

\noindent \textbf{Hardware Platform:} Our experiments was performed on a machine with 2 Intel(R) Xeon(R) Silver 4214R CPUs and 376GB DRAM, and 4 NVIDIA GeForce RTX 3090 GPUs, each with 24GB of HBM memory.

\noindent \textbf{Workload and Setting:} 
For DeepSpeed-Chat, we evaluated OPT~\cite{OPT}, where the Actor and Reference model are OPT-1.3b, and the Critic and Reward are OPT-350m model. For ColossalChat, we tested OPT with the same size. For GPT-2~\cite{GPT2}, the Actor and Reference are GPT2-xl, and the Critic and Reward are GPT2-medium. 

For each framework, we use its default input data, and we set the LoRA~\cite{LORA} dimension to 128.
We set the batch size to 2 for DeepSpeed-Chat, and 32 for ColossalChat. ColossalChat offloads the inference models to the CPU during actor and critic training.

\subsection{What Causes High Memory Consumption during RLHF?}
\label{sec:exp1}

\begin{figure}[htbp]
  \centering
  \includegraphics[width=0.95\linewidth]{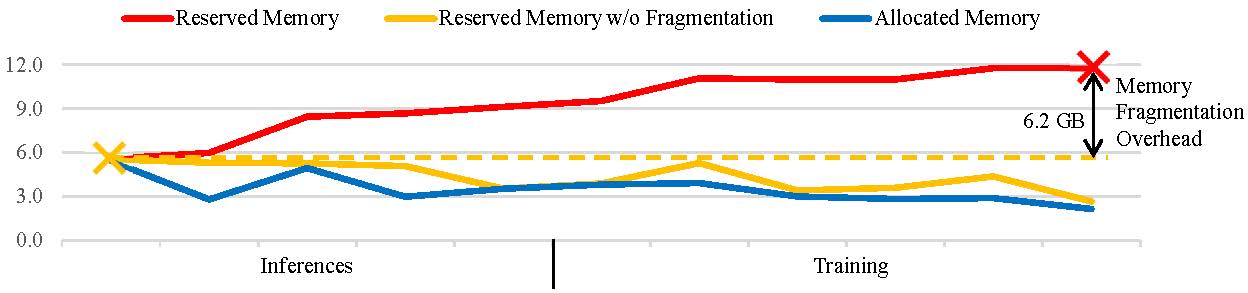}
  \caption{Memory usage (GB) of DeepSpeed-Chat running OPT with multiple memory management strategies enabled. The red cross marks the peak of reserved memory, while the yellow cross and dotted yellow line mark the theoretical peak of reserved memory after subtracting the size of memory fragmentation. }
  \label{fig:1}
\end{figure}

We ran DeepSpeed-Chat with ZeRO, CPU offloading, and gradient checkpointing enabled -- hereafter referred to as memory management strategies -- and profiled the memory usage as shown in Figure~\ref{fig:1}. In this figure, ``reserved memory w/o fragmentation'' is calculated by subtracting the size of fragmentation from the reserved memory (the yellow line). The difference between the peak reserved memory and ``reserved memory w/o fragmentation'' is referred to as \textbf{memory fragmentation overhead}.

In this experiment, the peak memory usage appears in the training phase where the reserved memory size is much larger than the allocated memory size. We further found that the gap between the reserved and allocated memory is due to external memory fragmentation (hereafter referred to as fragmentation).  In this example, the memory fragmentation overhead is 6.2 GB, increasing the memory consumption by 46\%. 

To determine whether significant fragmentation accumulates from the prior tasks, we compared the memory consumption of the three scenarios: (1) running both inferences and training; (2) training the actor and critic models with pre-collected data; (3) training only the actor model with pre-collected data. When only performing the training phases, we observed smaller fragmentation and reserved memory. Therefore, most fragmentation is accumulated from inferences to the training. In \S\ref{sec:exp3}, we further confirmed that the inferences generate the most fragmentation, which introduces most of the memory consumption to RLHF.

\begin{mdframed}
\textbf{Insight:} RLHF introduces high memory consumption because of memory fragmentation. The memory consumption reaches its peak during the training, but mostly due to a large size of memory fragmentation accumulated from inferences.
\end{mdframed}

\subsection{How Effective are Different Memory Management Strategies?}
\label{sec:exp2}

\begin{table*}[htbp]
\centering
\footnotesize
\caption{Memory usage under different memory management strategies. ``Reserved'' shows the peak of reserved memory size, ``Frag.'' column shows the size of memory fragmentation, and ``Allocated'' column lists the peak size of allocated memory. We highlighted the strategies (with red color) that increase the memory fragmentation overhead, and the cases (with bold format) where \texttt{empty\_cache()} is effective in reducing the fragmentation. \label{table:1}}
\scalebox{0.88}{
\begin{tabular}{|l|l|l|r|r|r|r|r|}
\hline
\multicolumn{1}{|l|}{\multirow{2}{*}{Framework}} & \multicolumn{1}{l|}{\multirow{2}{*}{Model}} & \multicolumn{1}{l|}{\multirow{2}{*}{Strategy}} & \multicolumn{2}{c|}{Original} & \multicolumn{1}{c|}{\multirow{2}{*}{Allocated}} & \multicolumn{2}{c|}{Using \texttt{empty\_cache()}} \\ \cline{4-5} \cline{7-8}
                                &                            &                              & Reserved & Frag. &  & Reserved & Frag.  \\ \hline
\multirow{7}{*}{DeepSpeed-Chat} & \multirow{6}{*}{OPT}       & None                   & 18.8  & 0.2   & 18.2    &     19.4     &      <0.1       \\ \cline{3-8} 
                                &                            & ZeRO-1                 & 15.6  & 0.1   & 14.4    &    15.9      &   0.1         \\ \cline{3-8} 
                                &                            & \textcolor{red}{ZeRO-2}& 14.5  & \textcolor{red}{0.6}   & 12.8    &     \textbf{14.3}     &     \textbf{<0.1}        \\ \cline{3-8} 
                                &                            & \textcolor{red}{ZeRO-3}& 17.3  & \textcolor{red}{3.7}   & 12.0    &    \textbf{13.7}      &   \textbf{0.3}          \\ \cline{3-8} 
                                &                            & \textcolor{red}{ZeRO-3 + CPU Offloading}         & 15.4   &  \textcolor{red}{4.0}  &  9.8  &     \textbf{11.7}     &    \textbf{0.3}        \\ \cline{3-8}
                                &                            & Gradient Checkpointing & 15.4  & 0.6   & 14.8    &     15.4     &    0.1         \\ \cline{3-8} 
                                &                            & \textcolor{red}{All Enabled} & 11.8  & \textcolor{red}{6.2}   &  5.4   &     \textbf{5.9}     &     \textbf{0.1}        \\ \hline
                                
\multirow{9}{*}{ColossalChat}   & \multirow{4}{*}{OPT}       & None                   & 17.5  & 0.2   & 17.0    &     17.8     &    0.4        \\ \cline{3-8} 
                                &                            & \textcolor{red}{ZeRO-3}& 16.5  & \textcolor{red}{0.5}   & 15.6    &     \textbf{16.4}     &   \textbf{0.4}          \\ \cline{3-8} 
                                &                            & ZeRO-3 + CPU Offloading                 &  13.1  &  0.4  &   12.3   &      13.1     &       0.2       \\ \cline{3-8}
                                &                            & \textcolor{red}{Gradient Checkpointng}                 &  14.8  & \textcolor{red}{0.7}   &  12.1    &    \textbf{12.5}       &   \textbf{0.1}           \\ \cline{2-8}

                                & \multirow{5}{*}{GPT-2}     & None                   & 22.9  & 6.9   & 14.0    &    \textbf{15.0}      &    \textbf{0.1}         \\ \cline{3-8} 
                                &                            & \textcolor{red}{ZeRO-3}&  22.1 &  \textcolor{red}{7.6}  &  13.2   &      \textbf{16.6}    &  \textbf{0.2} \\ \cline{3-8}
                                                                &                            & ZeRO-3 + CPU Offloading                 &  15.0 &  2.6  &   10.3  &   \textbf{11.5}     &  \textbf{0.1}           \\ \cline{3-8}
                                &                            & Gradient Checkpointing                 &  22.9 & 6.9  &   14.0  &   \textbf{15.0}       &    \textbf{0.1}           \\ \cline{3-8}
                                &                            & All Enabled                 & 15.0  & 2.6   &   10.3  &     \textbf{11.5}     &    \textbf{0.1}         \\ \hline
                                
\end{tabular}
}
\end{table*}

We profiled the memory usage of DeepSpeed-Chat and ColossalChat with different strategies enabled, as shown in Table~\ref{table:1}. Note that ColossalChat does not support ZeRO-1.
In addition, ColossalChat fails in the gradient synchronization when all strategies are enabled, so we excluded those cases. 
For each of the strategies, we have the following observations: 

\noindent \textbf{ZeRO-1:} Based on our investigation, ZeRO-1 does not increase the memory fragmentation overhead, and it stably reduces memory consumption (the reserved memory).

\noindent \textbf{ZeRO-2:} ZeRO-2 of DeepSpeed-Chat can slightly increase the fragmentation. However, it still reduces memory consumption.

\noindent \textbf{ZeRO-3:} ZeRO-3 increases the memory fragmentation overhead. In DeepSpeed-Chat, it causes the memory consumption even greater than those with ZeRO-1 and ZeRO-2, mostly due to memory fragmentation. 

\noindent \textbf{CPU Offloading:} Based on our observation, CPU offloading can affect the size of fragmentation. However, it can effectively reduce the memory consumption. 

\noindent \textbf{Gradient Checkpointing:} It may slightly increase the memory fragmentation overhead, but it is still effective in reducing memory consumption, except for ColossalChat with GPT-2. We further figured out that for GPT-2, the memory consumption reaches its peak during the inference phases, where gradient checkpointing has no effect.

\begin{mdframed}
\textbf{Insights:} 
Not all memory management strategies can reduce memory consumption: ZeRO-3 may increase the fragmentation and memory consumption; ZeRO-2 and CPU offloading may increase the fragmentation, but they still effectively reduce memory consumption; gradient checkpointing only reduces the memory consumption when peaks occur during training phases, and ZeRO-1 consistently lowers memory consumption.
\end{mdframed}

\subsection{How Can We Effectively Reduce Memory Consumption in RLHF? }
\label{sec:exp3}

We find that the \texttt{empty\_cache()} API 
exposed by the PyTorch allocator can help significantly alleviate memory fragmentation. 
Invocation of the API will release all cached memory blocks back to the GPU~\cite{EmptyCache}. 
To reduce memory consumption, we propose to insert \texttt{empty\_cache()} after each inference and training phase to release cached memory. 
Our results show that the proposed approach effectively reduces memory fragmentation, as shown in the bold part of Table~\ref{table:1}. For these cases, it reduces the memory consumption by 25\% on average. Additionally, the approach only increases the end-to-end time overhead by 2\% on average.

We compared the memory consumption of invoking \texttt{empty\_cache()} at different phases: (1) after each inference and training phase (2) only after each inference phase (3) only after the training phases. 
Based on our evaluation, invoking \texttt{empty\_cache()} after inferences is almost as effective as invoking upon each inference and training phase, while invoking only after the training phases is not very effective. 
The observation echoes our previous insight in \S\ref{sec:exp1} that the inference phase introduces most of the memory fragmentation and has the largest impact on RLHF's memory consumption.

\begin{mdframed}
\noindent \textbf{Insight:} Invoking \texttt{empty\_cache()} after each inference phase can significantly reduce the memory fragmentation overhead and memory consumption, with only increasing 2\% end-to-end time on average.

\end{mdframed}
\section{Conclusion}
RLHF is an important stage for LLM alignment, but it often has high memory consumption. This paper provides the first study on RLHF memory usage. We identified the cause of high memory consumption, and investigated the effectiveness of different memory management strategies in the RLHF scenario. We also found a simple yet effective approach to reduce memory consumption by using the \texttt{empty\_cache()} API, which can reduce 25\% of the memory consumption with only 2\% end-to-end time overhead on average.


\newpage
\appendix
\section{Appendix: PyTorch Allocator and empty\_cache()}

The PyTorch memory allocator manages a pool of free memory blocks to minimize the overhead associated with frequent allocations and deallocations directly from the GPU. When memory is requested, the allocator first checks this pool for suitable blocks, invoking the costly \texttt{cudaMalloc()} operation only when necessary. Upon deallocation, memory blocks are not immediately returned to the GPU but are instead cached for future reuse. In RLHF training, different tasks often have different memory allocation patterns with varying object sizes, leaving smaller memory blocks in the pool that are difficult to reuse, leading to memory fragmentation. When memory management strategies are used, the difference in allocation sizes can become even more noticeable, making the fragmentation worse.

The PyTorch API function, \texttt{empty\_cache()}, releases all unused cached blocks in the pool, reducing memory fragmentation. The function \texttt{empty\_cache()} is not commonly used in traditional training because it has a limited impact on memory savings and significant time overhead. However, it is effective in RLHF training, likely because \texttt{empty\_cache()} can release most of the cached blocks from the previous task, preventing fragmentation. Additionally, since the last task has already finished, these memory blocks are no longer being used by any stream, allowing them to be released without waiting.

\section{Appendix: Other Implementation Details}
We implemented a profiler that collects the sizes of reserved and allocated memory by calling the API of PyTorch's allocator. External memory fragmentation is computed at each \texttt{cudaMalloc()} invocation. It represents the difference between reserved and allocated memory when the allocator cannot satisfy the requested size due to non-contiguous freed objects.

For ColossalChat, we observed that the memory consumption of \texttt{generation()}  was exceptionally high in the original implementation. 
We replaced its implementation of the function with HuggingFace's to decrease memory consumption.

\section{Appendix: Additional Experiment Results}

\begin{table*}[htbp]
\centering
\footnotesize
\caption{Memory usage with and without ZeRO-3 on a node with 4 A100 GPUs. “Reserved” shows the peak of reserved memory size, “Frag.” column shows the size of memory fragmentation, and “Allocated” column lists the peak size of allocated memory. We highlighted the strategies (with red color) that increase the memory fragmentation overhead, and the cases (with bold format) where \texttt{empty\_cache()} is effective in reducing the fragmentation.} \label{table:2}
\scalebox{0.88}{
\begin{tabular}{|l|l|l|r|r|r|r|r|}
\hline
                               &                              & \multicolumn{1}{c|}{}                           & \multicolumn{2}{c|}{Original}                               & \multicolumn{1}{c|}{}                            & \multicolumn{2}{c|}{Using \texttt{empty\_cache()}}         \\ \cline{4-5} \cline{7-8} 
\multirow{-2}{*}{Framework}    & \multirow{-2}{*}{Model}      & \multicolumn{1}{c|}{\multirow{-2}{*}{Strategy}} & \multicolumn{1}{l|}{Reserved} & Frag.                       & \multicolumn{1}{c|}{\multirow{-2}{*}{Allocated}} & \multicolumn{1}{l|}{Reserved}      & Frag.        \\ \hline
                               &                              & None                                            & \multicolumn{1}{l|}{46.4}     & 2.4                         & 43.5                                             & \multicolumn{1}{l|}{\textbf{45.5}} & \textbf{0.3} \\ \cline{3-8} 
                               & \multirow{-2}{*}{OPT-1.3b}   & {\color[HTML]{FF0000} ZeRO-3}                   & \multicolumn{1}{l|}{46.4}     & {\color[HTML]{FF0000} 2.9}  & 43.2                                             & \multicolumn{1}{l|}{\textbf{45.0}}   & \textbf{0.3} \\ \cline{2-8} 
                               &                              & None                                            & \multicolumn{1}{l|}{53.4}     & 9.2                         & 31.4                                             & \multicolumn{1}{l|}{\textbf{44.3}} & \textbf{0.1} \\ \cline{3-8} 
                               & \multirow{-2}{*}{OPT-6.7b}   & {\color[HTML]{FF0000} ZeRO-3}                   & \multicolumn{1}{l|}{55.3}     & {\color[HTML]{FF0000} 20.6} & 25.6                                             & \multicolumn{1}{l|}{\textbf{50.3}} & \textbf{0.8} \\ \cline{2-8} 
                               &                              & None                                            & \multicolumn{1}{l|}{56.2}     & 8.8                         & 39.2                                             & \multicolumn{1}{l|}{\textbf{44.9}} & \textbf{0.2} \\ \cline{3-8} 
\multirow{-6}{*}{ColossalChat} & \multirow{-2}{*}{Llama-2-7b} & {\color[HTML]{FF0000} ZeRO-3}                   & \multicolumn{1}{l|}{60.5}     & {\color[HTML]{FF0000} 13.4} & 32.3                                             & \multicolumn{1}{l|}{\textbf{54.5}} & \textbf{1.7} \\ \hline
\end{tabular}
}
\end{table*}

We also tested more examples on machines with A100s, as shown in Table~\ref{table:2}. We found that the observations discussed in the main text also hold true across different platforms and models.

\section{Appendix: Limitations}
Our studies include just two open-sourced RLHF frameworks and two models. Our findings might vary when applied to other frameworks or models. Due to page constraints, we did not list the underlying reasons behind the varying effectiveness of each strategy. Additionally, we did not cover all existing memory management strategies, such as PagedAttention~\cite{PagedAttention}.

\end{document}